%% file: main.tex
\pdfoutput=1
\documentclass[letterpaper]{article} 

    \usepackage{aaai2027}

\input{commands}

\nocopyright

\title{\textsc{DeoGen}: Automatic Benchmark Generation for Reasoning over Policies}
\author{
    Sam Kouteili*\textsuperscript{\rm 1,2},
    William Liang*,
    Keeran Murray\textsuperscript{\rm 1},
    Adriana Zhou\textsuperscript{\rm 1},
    Ramesh Viswanathan\textsuperscript{\rm 3},
    Scott Shapiro\textsuperscript{\rm 3,2},
    Ruzica Piskac\textsuperscript{\rm 1,2}
}
\affiliations{
    \textsuperscript{\rm 1}Yale University\\
    \textsuperscript{\rm 2}Leibniz AI, Inc.\\
    \textsuperscript{\rm 3}Yale Law School\\
    sam.kouteili@yale.edu
}

\begin{document}

\maketitle

\begin{abstract}
Language models are increasingly being deployed in regulated settings, where a model must apply given rules to the facts of a situation to determine a verdict. Benchmarking this capability, known as \emph{deontic reasoning}, remains a costly and manual process, with existing datasets largely hand-crafted. In this paper, we present \tool{}, a scalable, automatic, and parametrizable benchmark generator for deontic reasoning. By autoformalizing policy rules to first-order logic, we can construct decision trees that capture possible outcome trajectories. Each path down a tree becomes a benchmark entry, with each tree producing combinatorially many entries relative to tree depth. Paths ending at a leaf entail a verdict, while non-leaf paths measure a model's ability to abstain when presented with insufficient information. 
Evaluations show frontier models return the correct verdict 89--92\% of the time. Notably, performance degrades (22--40\% correct) on entries without sufficient information to claim a verdict, even with explicit instructions to abstain in such cases.
In all, \tool~presents an automatic, scalable, and parametric method for generating deontic reasoning benchmarks. 
\end{abstract}

\input{sections/1-introduction}
\input{sections/3-prelim}

\input{sections/4-method}

\input{sections/5-setup}

\input{sections/6-evaluation}

\input{sections/2-related}

\input{sections/7-conclusion}

\bibliography{references}


\end{document}

%% file: commands.tex
\usepackage{times}  
\usepackage{helvet}  
\usepackage{courier}  
\usepackage[hyphens]{url}  
\usepackage{graphicx} 
\usepackage{natbib}  
\usepackage{caption} 
\usepackage{amsmath}
\usepackage{amssymb}
\usepackage{amsthm}
\theoremstyle{definition}
\newtheorem{definition}{Definition}

\newcommand{\tool}{\textsc{DeoGen}}

\usepackage{colortbl}
\usepackage{booktabs}
\usepackage{subcaption}
\usepackage{tikz}
\usetikzlibrary{calc,positioning,fit}

\definecolor{figblue}{HTML}{2B6CB0}
\definecolor{figgreen}{HTML}{2F855A}
\definecolor{figred}{HTML}{C53030}
\definecolor{figamber}{HTML}{B7791F}
\definecolor{figgray}{HTML}{4A5568}

\newcommand{\yeslab}{\textcolor{figgreen}{\textbf{Yes}}}
\newcommand{\nolab}{\textcolor{figred}{\textbf{No}}}
\newcommand{\neilab}{\textcolor{figamber}{\textbf{Not enough info}}}

\usepackage{xcolor}

\newsavebox{\promptboxsavebox}

\newcommand{\tinyish}{\fontsize{7}{7.5}\selectfont}


%% file: sections/1-introduction.tex
\section{Introduction}

\input{figures/pipeline}

Large Language Models (LLMs) are becoming ever-present, with adoption rising in highly-regulated domains such as financial compliance \citep{shin2023adoption}, insurance claims processing \citep{lieberthal2026claims}, and pharmaceuticals \citep{thirunavukarasu2023medicine}. While model reasoning has notably improved, hallucinations remain a problem, with errors in such high-stakes domains leading to potentially catastrophic outcomes: OpenAI models deployed to transcribe patient visits in hospitals were shown to fabricate medications and clinical details, corrupting patient records \citep{koenecke2024careless}. Moreover, recent rulings in New York have begun holding model users accountable for generated outputs \citep{mata2023avianca}, highlighting the dangers when the models do not properly apply the rules. 

To assess and train model capabilities to correctly apply the rules, we need benchmarks. 
Several datasets exist examining how models apply rules to situational facts \citep{sun2022conditionalqa,zhou2025rulearena,dou2026deonticbench}. These datasets broadly fit under the category of \emph{deontic reasoning} benchmarks. Deontic reasoning concerns reasoning about normative concepts such as obligations, permissions, and prohibitions. Specifically, it involves applying rules to factual situations to determine a verdict. In these benchmarks, models are typically asked to interpret a collection of rules, combine them with a factual scenario, and determine whether a particular action or outcome is compliant with the governing norms.

Existing benchmarks are typically hand-crafted or aggregated from a pool of cases. This creates two fundamental limitations. First, as models improve and data increasingly appears during training, these benchmarks saturate, with no real scalable means of creating new entries. Recent works demonstrate this trend, with agents achieving 99.2\% soundness on ConditionalQA \cite{bayless2025neurosymbolic}.
Second, benchmark construction and correctness are difficult to verify. Human annotation is expensive, time-consuming, and error-prone, while LLM-generated samples can also contain incorrect labels or subtle logical inconsistencies~\cite{gema2025mmlu}.

To address these limitations, we introduce \tool{}, an automatic benchmark
generator for deontic reasoning. 
In recent years, we have witnessed several examples of automated benchmark generation, primarily for domains with rich formal structure such as mathematics and code~\citep{saturn2025,shao2024deepseekmath}. In these settings, benchmarks can often be generated automatically as the problem is formally defined, and thus symbolic methods can enumerate reasoning tasks for both mathematical problem solving and code generation. In contrast, deontic reasoning lacks such a generator because its source material is unstructured, natural language rules, offering no symbolic handle for automatically generating benchmarks.
Autoformalization
presents a path forward: a growing body of work automatically translates natural-language statements into formal logic
\citep{bayless2025neurosymbolic, wu2022autoformalization,cosler2023nl2spec}. Once a policy
is autoformalized, we can systematically enumerate fact scenarios and formally deduce the verdict
each one entails, so ground-truth labels no longer need to be manually annotated.

Our tool leverages autoformalization to support benchmark enumeration. The example in Figure~\ref{fig:text-tree-table} provides a high-level illustration of the underlying algorithm. Given a simple natural language policy, a car is authorized to park if it displays a valid permit and is registered in the area,
we first autoformalize the rules to a symbolic representation:
{\small$ (\texttt{validPermit} \land \texttt{localRegistered}) \implies \texttt{authPark}$}. 
This formalization is then used to construct decision trees. Assuming that we are interested in the scenarios under which a car is authorized to park, Figure~\ref{fig:text-tree-table}b shows the corresponding decision tree. We efficiently generate benchmark scenarios by enumerating the paths in the decision tree: if {\small\texttt{validPermit}} and {\small\texttt{localRegistered}} are true, then {\small\texttt{authPark}} entails true; if {\small\texttt{validPermit}} is false, then {\small\texttt{authPark}} must be false. This way we can also consider scenarios where the verdict is not explicitly entailed. For example if we only know that the car has a valid permit ({\small \texttt{validPermit=true}}), we cannot assert whether it is authorized to park. This allows us to evaluate a model's ability to correctly abstain under insufficient information.  

\tool{} offers three key advantages over existing deontic reasoning benchmarks. \textbf{(1) Automatic}: Generating new benchmark entries simply requires selecting new paths through the decision tree. \textbf{(2) Scalable}:  Because the number of decision tree paths grows combinatorially with tree depth, a single source document can produce benchmark suites orders of magnitude larger than manually constructed counterparts. \textbf{(3) Parametric}: Benchmark instances can be generated by controlling the depth and diversity of the selected decision tree paths.

We run \tool{} on four policies: sections of the U.S. IRS tax code, an insurance cafeteria plan, a code of coverage, and a faculty hiring handbook, generating 451,047 benchmark entries.
We evaluate frontier (Claude Sonnet 5, GPT-5.5) and open-source (Qwen3.6, Gemma 4) models on a sample of 1,639 entries. On questions with determined verdicts, the models score 75.1--92.5\% correctness, showing a need to improve model deontic reasoning for high-stakes workflows. When the facts stop short of a verdict, so the correct answer is to abstain, the models collapse to 4.0--40.0\%. This is alarming in domains where not enough information is often a real outcome (imagine a layperson asking a chatbot if they are compliant with tax laws).
There exists a line of work investigating model abstention \cite{wen2025abstentionsurvey}, finding that on regulatory questions, sycophancy and non-abstention co-occur \cite{dahl2024legalfictions}. Our finding adds to this literature by not testing epistemic gaps (facts the model never knew), rather isolating \emph{logical underdetermination}: the policy is fully in context, the facts just don't entail a verdict.


%% file: figures/pipeline.tex
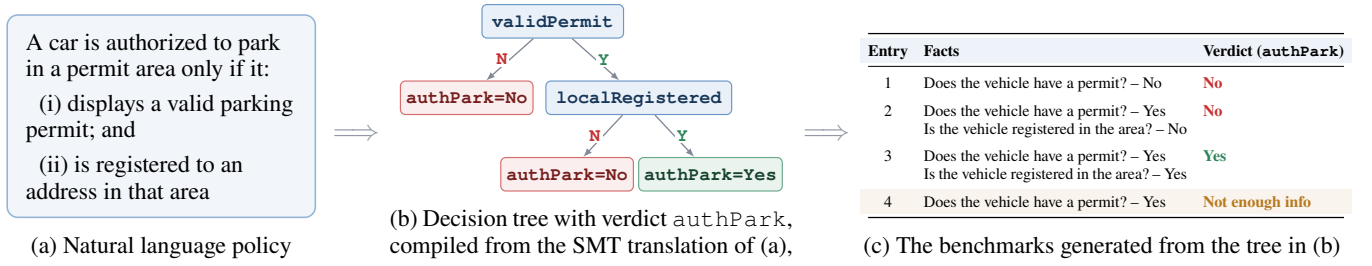
\begin{figure*}[t]
\centering

\begin{subfigure}[b]{0.24\textwidth}
\centering
\begin{tikzpicture}
\node[
    draw=figblue!50,
    fill=figblue!6,
    line width=0.6pt,
    rounded corners=4pt,
    inner sep=7pt,
    text width=0.84\linewidth,
    align=flush left,
    font=\small
] {
A car is authorized to park in a permit area only if it:\\[3pt]
\hspace*{5pt}(i) displays a valid parking permit; and\\[3pt]
\hspace*{5pt}(ii) is registered to an address in that area
};
\end{tikzpicture}
\caption{Natural language policy}
\end{subfigure}
\hfill
\raisebox{1.6cm}{\textcolor{figgray!60}{\large$\Longrightarrow$}}
\hfill
\begin{subfigure}[b]{0.30\textwidth}
\centering
\resizebox{0.95\linewidth}{!}{%
\begin{tikzpicture}[
    level distance=16mm,
    level 1/.style={sibling distance=36mm},
    level 2/.style={sibling distance=30mm},
    decision/.style={
        draw=figblue!70, fill=figblue!10, line width=0.7pt,
        rectangle, rounded corners=3pt,
        minimum height=7mm, inner xsep=6pt,
        font=\large\bfseries\ttfamily, text=figblue!50!black
    },
    yesleaf/.style={
        draw=figgreen!70, fill=figgreen!10, line width=0.7pt,
        rectangle, rounded corners=3pt,
        minimum width=17mm, minimum height=7mm, align=center,
        font=\large\bfseries\ttfamily, text=figgreen!50!black
    },
    noleaf/.style={
        draw=figred!70, fill=figred!8, line width=0.7pt,
        rectangle, rounded corners=3pt,
        minimum width=17mm, minimum height=7mm, align=center,
        font=\large\bfseries\ttfamily, text=figred!60!black
    },
    yeslabel/.style={fill=white, inner sep=1pt,
        font=\large\bfseries\ttfamily, text=figgreen},
    nolabel/.style={fill=white, inner sep=1pt,
        font=\large\bfseries\ttfamily, text=figred},
    edge from parent/.style={draw=figgray!70, line width=0.8pt, -latex}
]
\node[decision] {validPermit}
    child {
        node[noleaf] {authPark=No}
        edge from parent node[left,nolabel] {N}
    }
    child {
        node[decision] {localRegistered}
            child {
                node[noleaf] {authPark=No}
                edge from parent node[left,nolabel] {N}
            }
            child {
                node[yesleaf] {authPark=Yes}
                edge from parent node[right,yeslabel] {Y}
            }
        edge from parent node[right,yeslabel] {Y}
    };
\end{tikzpicture}%
}
\caption{Decision tree with verdict  \texttt{authPark}, compiled from the SMT translation of (a), }
\end{subfigure}
\hfill
\raisebox{1.6cm}{\textcolor{figgray!60}{\large$\Longrightarrow$}}
\hfill
\begin{subfigure}[b]{0.36\textwidth}
\centering
\resizebox{\linewidth}{!}{%
\renewcommand{\arraystretch}{1.4}
\setlength{\tabcolsep}{5pt}
\begin{tabular}{@{}c p{5.8cm} l@{}}
\toprule
\rowcolor{figblue!8}
\textbf{Entry} & \textbf{Facts} & \textbf{Verdict (\texttt{authPark})} \\
\midrule
1 & Does the vehicle have a permit? -- No & \nolab \\
\addlinespace[2pt]
2 & Does the vehicle have a permit? -- Yes \newline
    Is the vehicle registered in the area? -- No  & \nolab \\
\addlinespace[2pt]
3 & Does the vehicle have a permit? -- Yes \newline
    Is the vehicle registered in the area? -- Yes & \yeslab \\
\addlinespace[2pt]
\rowcolor{figamber!8}
4 & Does the vehicle have a permit? -- Yes & \neilab \\
\bottomrule
\end{tabular}%
}
\caption{The benchmarks generated from the tree in (b)}
\end{subfigure}

\caption{\tool~pipeline, illustrating transformation from text to a decision structure and its tabular representation.}
\label{fig:text-tree-table}
\end{figure*}

%% file: sections/3-prelim.tex
\section{Preliminaries}
\label{sec:prelim}

\subsection{Boolean Satisfiability}
\label{sec:prelim-sat}

Boolean satisfiability is the problem of deciding whether a logical formula can be made true. Its basic building blocks are \emph{atomic propositions}, variables that are either true or false, each standing for an elementary statement (e.g., \emph{the vehicle displays a permit}). When solving, each atomic proposition requires an \emph{assignment}, which fixes a truth value for every variable. A formula $\varphi$ is \emph{satisfiable} if some assignment makes it true, and a set of formulas $\Gamma$ \emph{entails} $\varphi$, written $\Gamma \models \varphi$, if every assignment satisfying all of $\Gamma$ also satisfies $\varphi$. We work with quantifier-free formulas built from atomic propositions using the Boolean connectives ($\neg$, $\wedge$, $\vee$, $\rightarrow$). 

\emph{Satisfiability Modulo Theories} (SMT) generalizes this by letting atoms be constraints interpreted by theories such as integer arithmetic, e.g., $\mathit{age} < 24$. We write formulas in the SMT-LIB standard \citep{barrett2017smtlib} and check them with the Z3 solver \citep{demoura2008z3}.

\subsection{Deontic Reasoning as Entailment}
\label{sec:prelim-task}

We model a \emph{policy} as a finite set $P$ of such formulas, one per clause of the source text. A policy typically governs several determinations, each corresponding to an atom of $P$. Any of them can serve as the \emph{verdict atom}, written $\mathsf{verdict}$. \tool{} constructs one decision tree per chosen verdict, so a single policy yields as many trees, and thus benchmark families, as it has determinations. All subsequent definitions are relative to a fixed choice of verdict atom. A \emph{fact pattern} is a finite set $F$ of facts stated over the atoms of $P$ (e.g., $\mathit{age} = 19$). We require $P \cup F$ to be consistent; the verdict of the policy on the facts is then well defined:
\begin{equation}
\label{eq:judgment}
\mathcal{J}(P, F) =
\begin{cases}
\textsf{Yes} & \text{if } P \cup F \models \mathsf{verdict},\\
\textsf{No} & \text{if } P \cup F \models \neg\mathsf{verdict},\\
\textsf{NEI} & \text{otherwise.}
\end{cases}
\end{equation}
The third case, \emph{not enough information} (NEI), arises whenever the facts leave relevant clauses undetermined. The task we benchmark presents a model with natural-language renderings of $P$ and $F$ and asks it to recover $\mathcal{J}(P, F)$. Entries with $\mathcal{J}(P, F) = \textsf{NEI}$ test a model's ability to abstain.

\subsection{Policy Decision Trees}
\label{sec:prelim-trees}

Not to be confused with CART learned classifiers~\cite{cart}, policy decision trees are compiled logical reformulations of a policy.

\begin{definition}[Policy decision tree]
\label{def:tree}
A \emph{policy decision tree} for $P$ is a finite rooted tree in which (i)~each internal node asks about one condition appearing in $P$; (ii)~the node's outgoing edges cover the possible answers to that condition, exactly one of which can hold (\textsf{Yes}/\textsf{No} for a Boolean condition, one edge per value otherwise); and (iii)~each leaf is labeled with a verdict, \textsf{Yes} or \textsf{No}.
\end{definition}

A \emph{path} $\pi$ runs from the root to a node (not necessarily a leaf); the answers along its edges form a fact pattern, which we denote $F(\pi)$. Two properties tie the tree $T$ to the policy:

\begin{definition}[Soundness and reducedness]
\label{def:sound}
$T$ is \emph{sound} for $P$ if every root-to-leaf path $\pi$ entails its leaf's label, i.e., $\mathcal{J}(P, F(\pi))$ equals the label. $T$ is \emph{reduced} for $P$ if every non-leaf path $\pi$ satisfies $\mathcal{J}(P, F(\pi)) = \textsf{NEI}$.
\end{definition}

In a sound and reduced tree, every path is a fact pattern whose ground-truth label is known. Paths reaching a leaf entail that leaf's verdict, and every shorter path yields NEI. 

%% file: sections/4-method.tex




\section{Methodology}
\label{sec:method}

\tool{} runs in three stages: the policy is formalized as SMT formulas (\S\ref{sec:method-formulas}), the formulas are compiled into a policy decision tree (\S\ref{sec:method-compile}), and each tree path is certified and rendered as a benchmark entry (\S\ref{sec:method-entries}).

\subsection{Autoformalizing Policies}
\label{sec:method-formulas}

\tool{} consumes an SMT formalization of the policy, generated by an autoformalizer. The formalization must carry two pieces of metadata for each clause: the source-text fragment it translates and that fragment's section label (e.g., \S129(c)(1)). In our experiments, the formalizer is a proprietary translator from policy text to SMT, with each specification additionally reviewed by hand. Everything downstream is automatic once the SMT is confirmed. 

\subsection{Compiling Policy Decision Trees}
\label{sec:method-compile}

Given SMT formula $P$ and a verdict atom $v$, we construct a policy decision tree.
Z3 first enumerates all assignments of the decision variables consistent with $P$. The tree is built top-down over this assignment set: each node branches on one variable, chosen greedily to minimize an estimated subtree depth.
A variable whose value cannot change the verdict is pruned.
Soundness and reducedness (Definition~\ref{def:sound}) are checked per path in \S\ref{sec:method-entries}. Two aspects need care.
\begin{itemize}
    \item \emph{Derived conditions.} A condition that $P$ defines in terms of other conditions, such as whether an expense in the AFL plan is ``eligible'' under IRS \S213(d), is resolved by a subtree over its defining conditions (fig. \ref{fig:compile}). When generating the benchmark entry as text, the facts of the subtree walk are noted, with a pre- and post-amble stating we are entering context regarding a subtree verdict.
    \item \emph{Numeric variables.} Assignments cannot range over integer or real domains. Instead, each numeric variable's domain is partitioned into the regions induced by the threshold atoms of $P$, and Z3 model enumeration selects one representative per region (see fig.~\ref{fig:compile}). Numerics related by a shared atom, such as $\mathit{ppy} = 65 - \mathit{taxpayerAge}$, are enumerated jointly so that no path asserts an impossible combination, and any condition whose variables are already fixed is evaluated by substitution. Every path is therefore consistent with its own numeric choices, ensuring consistency of $P \cup F(\pi)$ per Eq.~\ref{eq:judgment}. 
\end{itemize}




\input{figures/fig-compile}

\input{figures/table-stats}

\subsection{Generating Benchmark Entries}
\label{sec:method-entries}

With the policy decision tree, we can now conduct walks to collect benchmark entries, with the chosen edge labels the facts of the situation. Before committing a benchmark entry, however, the path needs to be certified for two conditions:
\begin{enumerate}
    \item \emph{Inconsistent facts}. Subtrees are compiled independently, so they can combine inconsistent facts. The qualifying-child clause of §152(c) and the qualifying-relative clause of §152(d) each ask, in their own fragment, how the person is related to the taxpayer; branched independently, a path can call the person a child in one answer and a sibling in the other.
    \item \emph{Unentailed labels}. The facts are consistent but do not force the recorded verdict, with multiple potential outcomes. For example, with university funding, $\geq 5$ years of research experience, and a PhD, the Yale Hiring Fellows handbook states that this person can be hired as an Associate Research Scholar, Research Scholar, and Postdoctoral Associate. The \texttt{appointmentType} tree, however, assumes Associate Research Scholar as the only correct entailment. 
\end{enumerate}

\tool{} checks for these two failure modes by certifying every path. A definite entry with fact pattern $F(\pi)$ and recorded label $\ell$ is released only if we prove the following two conditions, each relevant to the respective failure modes:
\begin{equation}
\label{eq:gate}
P \cup F(\pi) \not\models \bot
\quad\text{and}\quad
P \cup F(\pi) \models v = \ell,
\end{equation}
the second check discharged as the unsatisfiability of $P \cup F(\pi) \cup \{v \neq \ell\}$. An NEI entry is released only if Z3 finds two distinct values $a \neq b$ with both $P \cup F(\pi) \cup \{v = a\}$ and $P \cup F(\pi) \cup \{v = b\}$ satisfiable, which by Eq.~\ref{eq:judgment} is exactly $\mathcal{J}(P, F(\pi)) = \textsf{NEI}$.


Certified entries are rendered as a numbered fact list (Figure~\ref{fig:text-tree-table}c) using the metadata phrasings of \S\ref{sec:method-formulas}, and each entry carries the section labels of the provisions whose formulas involve its facts, allowing us to evaluate model citations. 


\subsection{Parametric Selection} 
Each benchmark entry is annotated with its verdict, its fact count, and its \emph{signature}: the set of variables questioned along its path (not including the answer). 
These annotations allow us to parametrize the distribution of our generated entries: 
\begin{itemize}
    \item \emph{Fact count}. The number of facts a model must integrate.
    \item \emph{Cut depth}. Sets what NEI entries withhold: a prefix cut near the root leaves most of the determination open, while a cut just above a leaf omits a single fact.
    \item  \emph{The signature} defines path uniqueness: entries sharing a signature pair ask the same questions, differing only in answers or numeric values along the way, so they are variants of one class. Keeping at most $k$ entries per class yields structurally distinct benchmark sets.
\end{itemize}

%% file: figures/fig-compile.tex
\begin{figure}[t]
\centering
\resizebox{\columnwidth}{!}{%
\begin{tikzpicture}[
  decision/.style={draw=figblue!70, fill=figblue!10, line width=0.7pt,
      rectangle, rounded corners=3pt, minimum height=6.5mm, inner xsep=5pt,
      align=center, font=\small, text=figblue!50!black},
  yesleaf/.style={draw=figgreen!70, fill=figgreen!10, line width=0.7pt,
      rectangle, rounded corners=3pt, minimum width=9mm, minimum height=6mm,
      align=center, font=\small\bfseries, text=figgreen!50!black},
  noleaf/.style={draw=figred!70, fill=figred!8, line width=0.7pt,
      rectangle, rounded corners=3pt, minimum width=9mm, minimum height=6mm,
      align=center, font=\small\bfseries, text=figred!60!black},
  dots/.style={font=\small, text=figgray},
  yl/.style={fill=white, inner sep=1pt, font=\footnotesize\bfseries\ttfamily,
      text=figgreen},
  nl/.style={fill=white, inner sep=1pt, font=\footnotesize\bfseries\ttfamily,
      text=figred},
  el/.style={fill=white, inner sep=1pt, font=\footnotesize\itshape,
      text=figgray},
  arr/.style={draw=figgray!70, line width=0.8pt, -latex}
]
\node[decision] (a) at (14mm,0) {What type of insurance?};
\node[dots, below=6mm of a, xshift=-20mm] (other) {$\cdots$};
\draw[arr] (a) -- (other);
\node[font=\small\bfseries, align=center, text=figamber!70!black]
      (t) at (34mm,-19mm) {Eligible long-term-care\\premium (\S213(d)(10))?};
\begin{scope}[shift={($(t.south)+(-4mm,-7mm)$)}, scale=0.55, transform shape]
\node[decision] (c1) at (0,0) {Annual premium $\leq \$200$?};
\node[yesleaf, below=8mm of c1, xshift=-15mm] (l1) {Yes};
\node[decision, below=9mm of c1, xshift=13mm] (c2)
      {Annual premium $\leq \$2{,}500$?};
\node[noleaf, below=8mm of c2, xshift=-16mm] (l2) {No};
\node[decision, below=9mm of c2, xshift=13mm] (c3) {Age $\leq 40$?};
\node[noleaf, below=8mm of c3, xshift=-11mm] (l3) {No};
\node[dots, below=9mm of c3, xshift=8mm] (d) {$\vdots$};
\draw[arr] (c1) -- node[yl, left=2pt] {Y} (l1);
\draw[arr] (c1) -- node[nl, right=2pt] {N} (c2);
\draw[arr] (c2) -- node[nl, left=2pt] {N} (l2);
\draw[arr] (c2) -- node[yl, right=2pt] {Y} (c3);
\draw[arr] (c3) -- node[yl, left=2pt] {Y} (l3);
\draw[arr] (c3) -- node[nl, right=2pt] {N} (d);
\end{scope}
\node[draw=figamber!70, dashed, line width=0.7pt, rounded corners=4pt,
      inner sep=3mm, fit=(t)(c1)(l1)(l2)(l3)(c3)(d)] (box) {};
\draw[arr] (a.south) -- node[el, right=3pt, pos=0.45]
      {Qualified long-term care contract} (box.north);
\node[draw=figgray!50, fill=figgray!8, line width=0.6pt, rounded corners=3pt,
      inner sep=5pt, align=left,
      font=\tiny\ttfamily, anchor=north east]
      (smt) at ($(box.north west)+(5mm,-12mm)$) {%
(assert (= isEligibleLongTermCarePremium\\
\hspace*{4pt}(ite (<= taxpayerAge 40)\\
\hspace*{9pt}(<= premiumPaidLongTermCareInsurance 200)\\
\hspace*{4pt}(ite (<= taxpayerAge 50)\\
\hspace*{9pt}(<= premiumPaidLongTermCareInsurance 375)\\
\hspace*{4pt}(ite (<= taxpayerAge 60)\\
\hspace*{9pt}(<= premiumPaidLongTermCareInsurance 750)\\
\hspace*{4pt}(ite (<= taxpayerAge 70)\\
\hspace*{9pt}(<= premiumPaidLongTermCareInsurance 2000)\\
\hspace*{9pt}(<= premiumPaidLongTermCareInsurance 2500)))))))};
\end{tikzpicture}%
}
\caption{Fragment of the AFL medical-expense tree. The plan document admits long-term-care insurance premiums only when they are eligible under the IRS Code, so the qualified-contract branch expands directly into a subtree compiled from \S213(d)(10). Numeric values are handled by constructing trees that segment the number bins.
}
\label{fig:compile}
\end{figure}
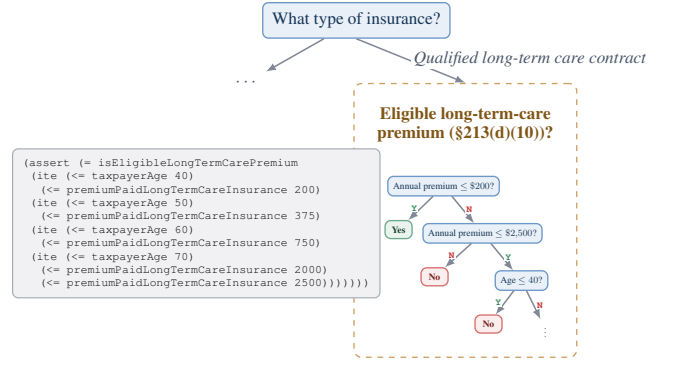

%% file: figures/table-stats.tex

\begin{table*}[t]
\centering
\tinyish
\setlength{\tabcolsep}{4pt}%
\newcommand{\pred}[1]{{\tiny\texttt{(#1)}} }%
\begin{tabular}{p{9.2cm}rrrrr}
\toprule
& & \multicolumn{2}{c}{Total possible entries} & \multicolumn{2}{c}{Evaluation set} \\
\cmidrule(lr){3-4} \cmidrule(lr){5-6}
Verdict Questions & Tree Depth & Def. & NEI & Def. & NEI \\
\midrule

\multicolumn{6}{l}{\textbf{IRS Code} (\S\S105(b), 125, 129, 152, 213; 9,156 words, 109 assertions)} \\
Is this individual a ``dependent'' of the taxpayer under \S152?  & 25 & 2,125 & 2,048 & 100 & 100 \\
\pred{isDependent}\\
Is the person a qualifying child, as defined in \S152(c)?  & 16 & 144 & 156 & 61 & 54 \\
\pred{isQualifyingChild}\\
Is this person a qualifying relative, as defined in \S152(d)? & 16 & 355 & 311 & 96 & 63 \\
\pred{isQualifyingRelative}\\
Does this expense qualify as ``medical care'' under \S213(d)?  & 12 & 56 & 40 & 45 & 30 \\
\pred{isMedicalCare}\\
Is this expense allowed as a deduction under \S213(a)?  & 40 & 59,922 & 61,937 & 100 & 100 \\
\pred{isDeductible}\\
Is this expense reimbursable from a Health FSA (excludable from gross income under \S105(b))? & 36 & 12,308 & 13,445 & 100 & 100 \\
\pred{isMedicalExcludedGrossIncome} \\
Is this a dependent care assistance program as defined in \S129(d)?  & 7 & 7 & 9 & 7 & 8 \\
\pred{isProgramDescribedInDDCA} \\
\midrule
\multicolumn{6}{l}{\textbf{AFL Plan Document} (14,527 words, 238 assertions)} \\
Is this an eligible medical expense as defined in Section 8.04? & 44 &  163,870 & 78,193 & 100 & 100 \\
\pred{EligibleMedicalExpense\_8\_04} \\
Is this an eligible dependent care expense as defined in Section 9.04(c)?  & 32 &  12,547 & 15,612 & 100 & 100 \\
\pred{EligibleDependentCareExpense\_9\_04} \\
Is this person a dependent for purposes of the Dependent Care Reimbursement Plan (Section 9.04(a))? \pred{isDependentUnderSectionIX} & 27 &  12,420 & 15,465 & 100 & 100 \\
\midrule
\multicolumn{6}{l}{\textbf{AFL Certificate of Coverage} (15,051 words, 1,190 assertions)} \\
Is this person a dependent? \pred{dependentInsured} & 9 & 33 & 31 & 33 & 29 \\
\midrule
\multicolumn{6}{l}{\textbf{Yale Hiring Fellows Policy} (2,828 words, 11 assertions)} \\
What is the type of appointment? \pred{appointmentType} & 5 & 8 & 5 & 8 & 5 \\
\midrule
\textbf{Total} &  & 263,795 & 187,252 & 850& 789 \\
\bottomrule
\end{tabular}
\caption{Benchmark statistics. For each determination: maximum decision-tree depth; the population of possible benchmark entries: definite entries (root-to-leaf paths) and NEI entries (root-to-internal-node paths). For the evaluation set, we draw up to 100 verdict-reaching + 100 NEI determinations, unless generated trees do not contain so many paths. 
}
\label{tab:stats}
\end{table*}

%% file: sections/5-setup.tex
\section{Experimental Setup}
\label{sec:setup}

\input{figures/table-main}

We generate benchmarks from four policies:
\begin{itemize}
    \item \emph{IRS Code} (\S\S105(b), 125, 129, 152, 213). Statutory text governing cafeteria benefit plans. Its definitions and provisions chain across sections. Whether an expense is reimbursable under \S105(b) turns on whether it is medical care under \S213(d), and the exclusions of \S\S105(b) and 129 both turn on who counts as a dependent under \S152.
    \item \emph{AFL Plan Document}. A cafeteria plan document from insurance carrier American Fidelity that lets employees pick among benefits options. The plan is explicitly contingent on the IRS Code, so its eligibility determinations incorporate the IRS sections by reference.
    \item \emph{AFL Certificate of Coverage}. An insurance certificate from the same carrier, describing what specific insurance policies actually cover.
    \item \emph{Yale Hiring Fellows Policy}. A university hiring policy whose verdict is an appointment type drawn from an enumeration rather than a Yes/No answer. 
\end{itemize}

Any variable in the formalization can serve as a verdict atom. We take as verdicts twelve determinations across the four policies the source texts themselves pose as questions. These determinations call one another as the sections do: dependent (\S152) resolves through qualifying child (\S152(c)) and qualifying relative (\S152(d)); the health-FSA exclusion (\S105(b)) opens both medical care (\S213(d)) and dependent (\S152); the dependent-care exclusion (\S129) invokes \S152. The AFL plan document is predicated on the IRS Code: eligible medical expense opens \S105(b) and \S213(d), and the dependent-care determinations open \S129 and \S152.

Table~\ref{tab:stats} reports, per verdict atom, the tree depth and the \emph{certified} populations of definite and NEI entries, pruned as per the two reductions of \S\ref{sec:method-entries}. In all, \tool{} generated 263,795 definite and 187,252 NEI entries, post-certification. On the plan-document medical-expense tree pruning eliminated 31,460 (question-set, verdict) classes with jointly inconsistent facts. On Fellows, three leaf paths were pruned whose eligibility criteria entail several appointment types at once.

From each population, we draw up to 100 definite and 100 NEI entries, sampling across shown-fact-count buckets (1--10, 11--20, 21--30, 31--40). When a population is smaller than the target, the evaluation set is the full certified population. For large trees we additionally cap structural repetition, choosing at most one entry per signature class, where a signature is the set of questions a path asks, so entries asking the same questions toward the same verdict, differing only in the answers such as numeric values along the way, appear once. On the health-FSA tree (verdict {\small\texttt{isMedicalExcludedGrossIncome}}) this cap removes 66\% of certified paths as variants of a path kept.

We grade by asking models to choose the verdict from a stated set of admissible answers, with ``Not enough information'' always among them. For the IRS and plan-document policies, each entry also carries the section labels of the provisions its facts touch (\S\ref{sec:method-formulas}); we grade cited provisions recall against these labels, crediting a citation of an ancestor provision at half weight, and report precision alongside. The certificate and hiring policies have no internal section labels, so citation grading covers the first two policies.

The frontier models (Claude Sonnet 5, GPT-5.5) are queried through their public APIs with extended reasoning disabled; the open-weight models are served with vLLM on a single NVIDIA H200 GPU of a university compute cluster and decoded greedily (temperature 0).

\paragraph{Manual Amendments.} Upon autoformalizing the source policies, we found that certain benchmark clauses were logically correct, but marginally disjoint from other clauses. For example, in AFL clauses 152(c)(1)(B) and 9.04(c), both covered the abode of an individual. “Same principal place of abode as the taxpayer for more than one-half of the taxable year” and “same principal place of abode as the Participant for more than half of the taxable year.” The autoformalization logic was correct; however, these two sections were never linked. We therefore manually add to the SMT formalizations bridges between variables that are assumed by the source policy. During manual review, we identified 33 distinct bridge asserts in the IRS and AFL formalizations.

%% file: figures/table-main.tex
\begin{table*}[t]
\centering
\small
\begin{tabular}{lcccccccc}
\toprule
& \multicolumn{2}{c}{IRS Code} & \multicolumn{2}{c}{AFL} & \multicolumn{2}{c}{Fellows} & \multicolumn{2}{c}{Average} \\
\cmidrule(lr){2-3} \cmidrule(lr){4-5} \cmidrule(lr){6-7} \cmidrule(lr){8-9}
Model & Def. & NEI & Def. & NEI & Def. & NEI & Def. & NEI \\
\midrule
Claude Sonnet 5      & 91.0 & 32.1 & 93.7 & 9.4 & 100.0 & 60.0 & 92.1 & 22.8 \\
GPT-5.5       & 88.3 & 55.8 & 91.3 & 19.1 & 62.5 & 80.0  & 89.2 & 40.0 \\
Qwen3.6-35B-A3B     & 83.9 & 5.9 & 92.2 & 1.5 & 87.5 & 0.0 & 87.2 & 4.0 \\
Gemma 4 31B         & 90.6 & 34.5 & 95.5 & 15.2 & 87.5 & 80.0 & 92.5 & 26.7 \\
Gemma 4 E4B         & 71.3 & 22.9 & 80.5 & 11.6 & 87.5 & 20.0 & 75.1 & 18.2 \\
\bottomrule
\end{tabular}
\caption{Accuracy (\%) on definite scenarios (Def.) and on underdetermined scenarios (NEI), whose keyed answer is \textsf{Not enough information}, per policy and averaged over all entries (AFL aggregates the plan document and certificate). Zero-shot; deterministic grading. Binomial 95\% CIs are at most $\pm3.5$ points; every definite--NEI gap is significant ($p{<}0.001$, two-proportion $z$-tests), the Gemma 4 31B--Claude definite gap is not.}
\label{tab:main}
\end{table*}

%% file: sections/6-evaluation.tex
\section{Evaluation}
\label{sec:eval}

We evaluate two frontier (Claude Sonnet 5, GPT-5.5) and three open-source (Gemma 4 31B, Gemma 4 E4B, Qwen3.6-35B-A3B) models on our generated dataset of 1,639 entries spanning the four source policy texts. Table~\ref{tab:main} displays the primary results.

On definite scenarios, the models cluster in a band that is competent but arguably insufficient for high-stakes work: four of the five score an average between 87.2\% and 92.5\%, with the small Gemma 4 E4B trailing at 75.1\%. Notably, the best model in the column is not a frontier model, but open-weight Gemma 4 31B (92.5\%), beating Claude Sonnet 5 (92.1\%) and GPT-5.5 (89.2\%). On underdetermined scenarios, performance is markedly worse: averages collapse from 87.2--92.5\% to 4.0--40.0\%. GPT-5.5 is the strongest abstainer at 40.0\%. Claude Sonnet 5, the stronger frontier model on definite entries, drops to 22.8\%, roughly half GPT-5.5's rate and below Gemma 4 31B (26.7\%). Qwen3.6-35B-A3B barely abstains at all (4.0\% overall and 0.0\% on the Fellows), committing a verdict in nearly all scenarios. 



\input{figures/fig-facts}

\input{figures/fig-abstention}

\paragraph{Accuracy by fact count.}  Figure~\ref{fig:facts} evaluates model performance over the number of facts presented. Across both determined and non-determined \textsf{NEI} verdicts, we see model performance generally decrease with longer chains. This decrease is more pronounced for \textsf{NEI} verdicts--we posit this happens due to the models assuming that more facts more likely implies a complete scenario. Qualitatively analyzing a sample of 50 output traces per model, we find that at longer depth, models still reason down the provisions and identify the conditions the facts never address, but now satisfy them by default: Claude Sonnet 5 on a length 38 AFL entry, told only that a person has not provided half of her own support, concludes this ``implies the taxpayer likely provides the support'' and Gemma 4 31B lists the residency and AGI facts for \S152(c)(4), notes they are absent, and commits regardless.

\paragraph{Failure modes.} We further analyze failure modes across definite and underdetermined scenarios. Figure~\ref{fig:abstention} shows that across definite scenarios, every model is more likely to claim an incorrect verdict than to abstain spuriously. In the underdetermined case, four of the five models are markedly more likely to claim ``yes'' than ``no'' on the binary verdicts of IRS and AFL. Among wrong commitments, 72--83\% answer Yes, with GPT-5.5 the outlier answering No on 56\% of its commitments: where the other models treat an unaddressed condition as satisfied, GPT-5.5 treats it as failed. The abstention trends of Table~\ref{tab:main} are also reflected here, in that models that perform worse on \textsf{NEI} entries, such as Claude Sonnet 5 and Qwen3.6, generally abstain less on definite scenarios too. This suggests that abstention behavior is tied to certain models with a higher or lower propensity to do so.

\paragraph{Citations.} Alongside verdicts, we also asked models to cite the relevant clauses. Table~\ref{tab:citations} presents our findings, showing that models cite precisely but not completely. What models cite is almost always relevant (precision 0.82--0.94), yet even the best model surfaces under a third of the provisions its verdict depends on (recall 0.293 for Claude Sonnet 5, falling to 0.162 for Gemma 4 E4B). Models that perform weaker exhibit higher precision but lower recall, suggesting that weaker models retreat to a few safe, obviously relevant citations.
This is consistent with \textsf{NEI} findings, as models that do not visit the provisions a determination depends on cannot notice that an unvisited branch is unresolved.

\input{figures/table-citations}

\paragraph{Few-shot prompting.} As an ablation, we apply few-shot prompting, where we show the model a sample policy and all the possible verdict paths. Specifically, we present the example benchmark entries from Figure~\ref{fig:text-tree-table}c. Despite always listing all the possible verdicts, including \textsf{NEI}, we wanted to examine whether explicitly demonstrating desired behavior, in particular for \textsf{NEI} cases, enabled models to correctly reason. Table~\ref{tab:prompting} shows it helps, but only partially. For three of the five models, NEI accuracy rises by 10--13 points (Claude Sonnet 5 22.8\% to 35.7\%, GPT-5.5 40.0\% to 51.1\%, Gemma 4 31B 26.7\% to 36.4\%), with no significant change in definite verdicts. Here, the best few-shot abstainer still misses half of its NEI entries. While in-context examples may shift a model's willingness to abstain, there remains a broader inability to track the provisions a verdict depends on.

\input{figures/table-prompting}

\paragraph{Human evaluation.} When given a representative sample of benchmark entries spanning all four source policy texts, two non-expert (undergraduates with no special legal training) human annotators achieved only $32\%$ accuracy, as shown in Table~\ref{tab:validity}. Inter-annotator agreement was also low at $52\%$. The human annotators cited unfamiliarity with legal code and source policy as main bottlenecks for accuracy. Underdetermined-case errors consisted of unwarranted commitment to Yes $45\%$ of the time and to No $25\%$ of the time. Correct abstention was achieved in $30\%$ of scenarios, which is an overall similar split to Claude Sonnet 5 and Gemma 4 31B. This suggests that both human annotators and models are biased towards Yes when committing incorrectly.

\paragraph{Closed-world formalization.} In their evaluation, human annotators raised issues with some benchmark entries which used implausible ages, such as
cases where a dependent was not of legal age but had a
spouse, stating that it was illegal for an underage person to be married. There were also benchmark entries where a dependent could be both the individual's spouse and their child. While these scenarios are improbable in the real world, they are logically permissible per the source document. The autoformalization-derived answers are faithful to only the source policy, whereas human annotators brought in normative assumptions.

 We assume a closed world environment and do not prohibit such cases during the generation process, attempting to isolate deontic reasoning and policy adherence. In our model prompt, we tell the models explicitly to only consider the source texts and to not assume external judgment. We choose to not encode normative world assumptions into the policy formalizations as additional bridges because we felt incorporating other intuitions without formalizing the additional laws themselves violates consistency. Our autoformalization thus remains unaffected by normative assumptions.
 

 \input{figures/table-validity}

\input{figures/fig-atypical.tex}

In Figure~\ref{fig:atypical}, we ablate for such situations. We segment entries that considered age determinations into real-world typical and atypical scenarios, where we deem whether or not a scenario is realistically plausible. $15\%$ of benchmark entries in IRS and AFL that involved age information fell into the  atypical age category. The figure shows that across all models, accuracy decreased in atypical age scenarios, with a 2.3--17.1 percentage-point loss. This is despite the fact that every model, when faced with atypical age facts, often noted them in their response and explicitly stated they would ignore it. GPT-5.5, which did the best in typical age scenarios ($70.7\%$ accuracy, compared to $63.2\%$ for Gemma 4 31B, the second-best) experienced the largest decrease of $17.1$ points in atypical age scenario accuracy.  We posit that this decrease in accuracy is due to inherent normative world assumptions the models hold, and find this an insightful metric to recognize model propensity to ground in prior knowledge of the world.



%% file: figures/fig-facts.tex
\begin{figure}[t]
\centering
\includegraphics[width=0.8\columnwidth]{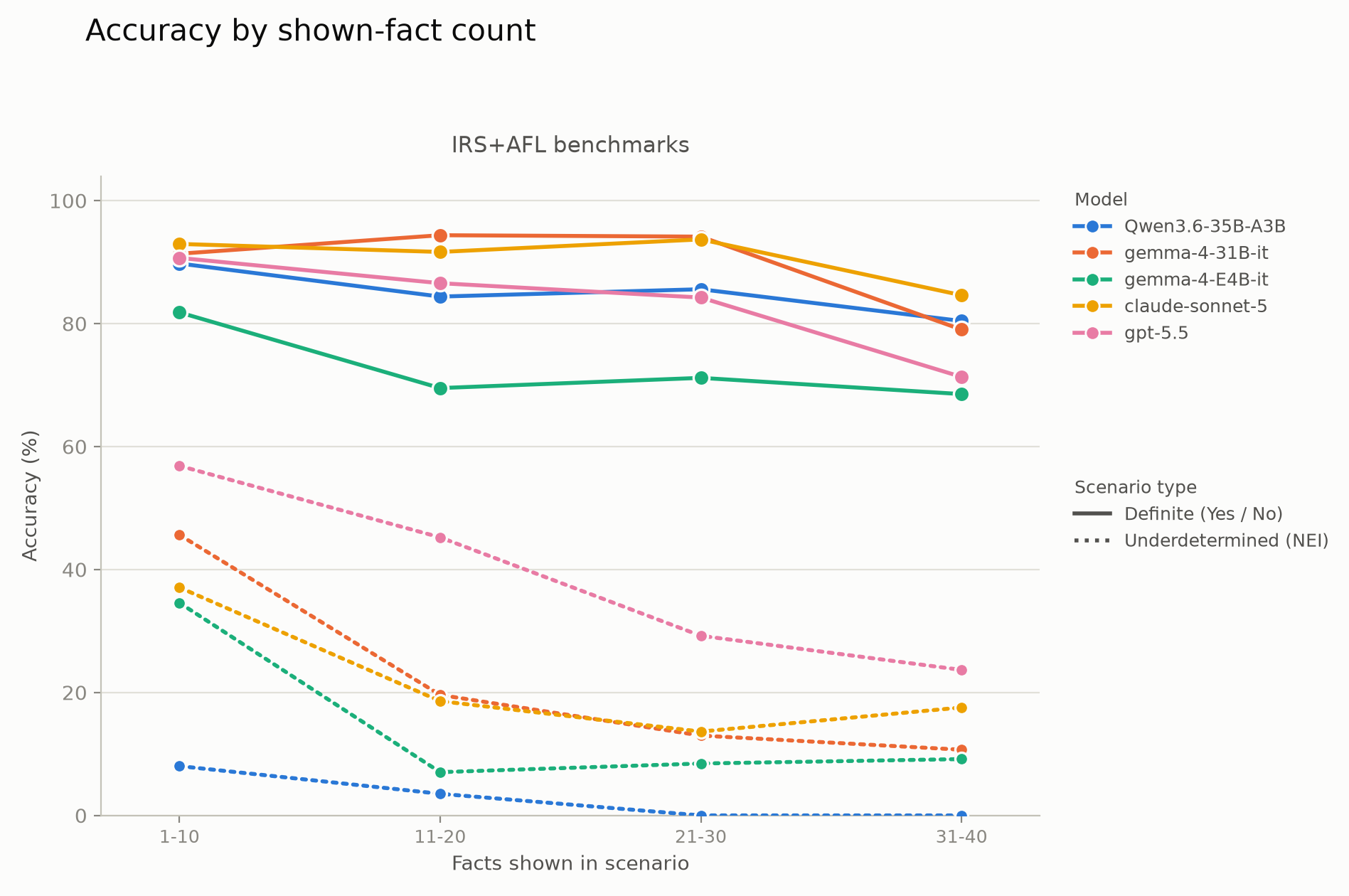}
\caption{Accuracy as a function of the number of facts shown. Path depth acts as a difficulty dial.}
\label{fig:facts}
\end{figure}

%% file: figures/fig-abstention.tex
\begin{figure}[t]
\centering
\includegraphics[
    width=0.92\columnwidth
]{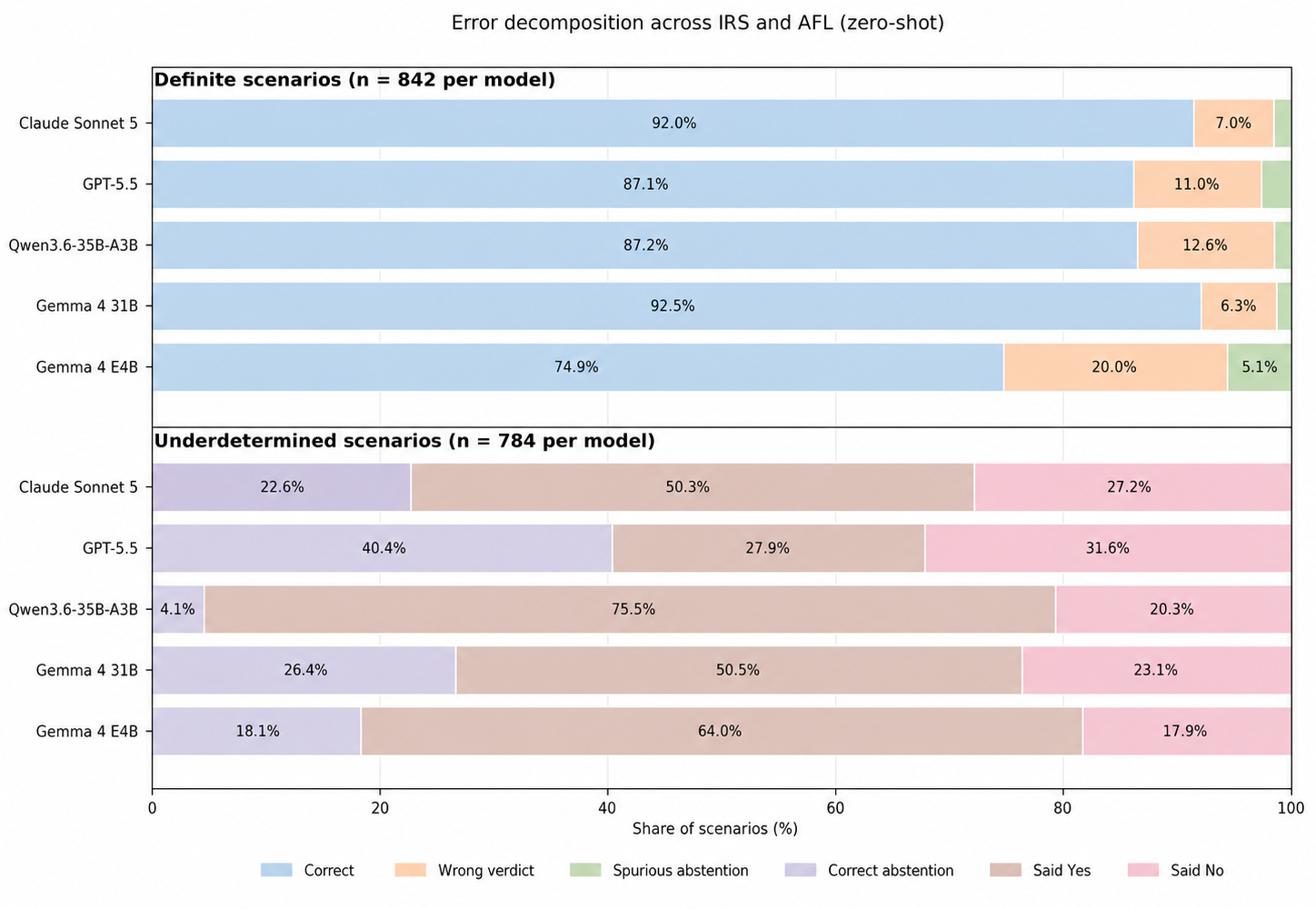}
\caption{
Error decomposition across zero-shot evaluations on the binary-verdict policies (IRS and AFL; Fellows excluded).
Bars show the distribution of outcomes within definite ($n=842$ per model) and underdetermined ($n=784$ per model) scenarios.
Definite-case errors are divided into wrong verdicts and spurious
abstentions; underdetermined-case errors are unwarranted commitments
to a verdict. One Qwen3.6 \textsf{NEI} response was unparsed.
}
\label{fig:abstention}
\end{figure}

%% file: figures/table-citations.tex
\begin{table}[t]
\centering
\small
\begin{tabular}{lcc}
\toprule
Model & Recall & Precision \\
\midrule
Claude Sonnet 5      & 0.293 & 0.824 \\
GPT-5.5       & 0.257 & 0.852 \\
Qwen3.6-35B-A3B     & 0.194 & 0.935 \\
Gemma 4 31B         & 0.194 & 0.884 \\
Gemma 4 E4B         & 0.162 & 0.912 \\
\bottomrule
\end{tabular}
\caption{Citation grounding against the provision-level gold for AFL and IRS: recall (did the model cite every provision its scenario depends on) and precision (were the cited provisions correct).}
\label{tab:citations}
\end{table}

%% file: figures/table-prompting.tex
\begin{table}[t]
\centering
\small
\begin{tabular}{lcccc}
\toprule
& \multicolumn{2}{c}{Zero-shot} & \multicolumn{2}{c}{Few-shot} \\
\cmidrule(lr){2-3} \cmidrule(lr){4-5}
Model & Def. & NEI & Def. & NEI \\
\midrule
Claude Sonnet 5      & 92.1 & 22.8 & 91.4 & 35.7 \\
GPT-5.5       & 89.2 & 40.0 & 87.1 & 51.1 \\
Qwen3.6-35B-A3B     & 87.2 & 4.0 & 90.0 & 1.5 \\
Gemma 4 31B         & 92.5 & 26.7 & 92.5 & 36.4 \\
Gemma 4 E4B         & 75.1 & 18.2 & 74.0 & 22.3 \\
\bottomrule
\end{tabular}
\caption{Few-shot ablation, averaged over all entries: worked examples including an \textsf{NEI} demonstration are drawn from a held-out determination. Demonstrating abstentions generally improves \textsf{NEI} performance; the 10--13-point gains are significant ($p{<}0.001$).}
\label{tab:prompting}
\end{table}

%% file: figures/table-validity.tex
\begin{table}[t]
\centering
\small
\begin{tabular}{lr}
\toprule
Measure & Result \\
\midrule
Human--gold label agreement ($n$ = 50, 2 annotators) & 32\,\% \\
NL rendering judged faithful to facts & 92\,\% \\
Inter-annotator agreement & 52\,\% \\
\bottomrule
\end{tabular}
\caption{Human validity study on a stratified sample. Annotators answer from the natural-language scenario and policy alone; they struggle to recover the gold label, yet judge the renderings faithful.}
\label{tab:validity}
\end{table}

%% file: figures/fig-atypical.tex
\begin{figure}[t]
\newcommand{\pred}[1]{{\texttt{#1}} }%
\centering
\includegraphics[
    width=0.90\columnwidth
]{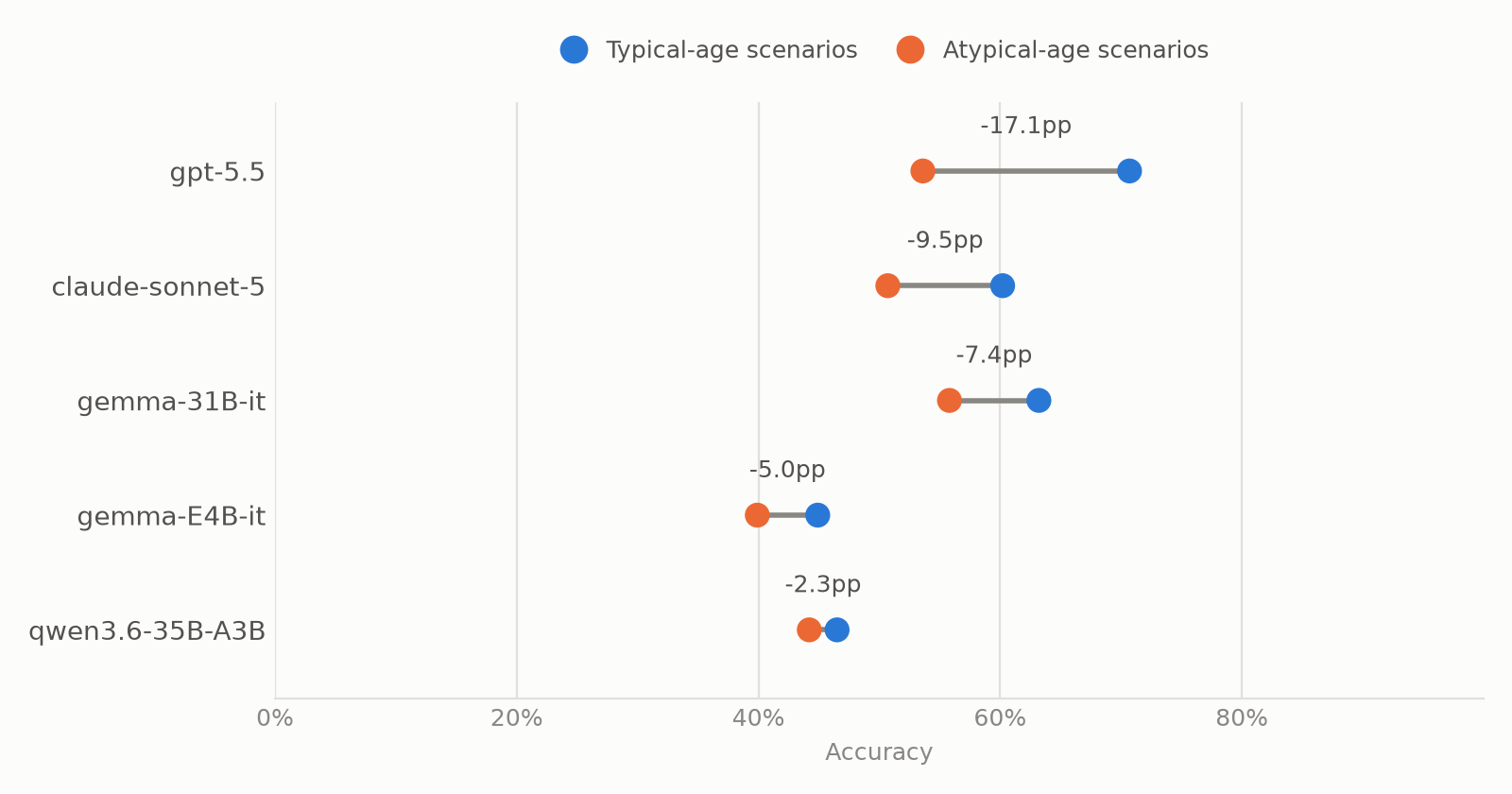}
\caption{
Accuracy across zero-shot scenarios in AFL and IRS which included age determinations (three IRS, two AFL).
Every model performs worse on atypical-age scenarios ($n=138$) than typical ones ($n=777$).
}
\label{fig:atypical}
\end{figure}

%% file: sections/2-related.tex
\section{Related Work}

\paragraph{Law and logic.}
There exists a long tradition of logic-based legal modeling \citep{prakken2015law}:
Four decades ago, the British Nationality Act was implemented as an executable logic program \citep{sergot1986british}. Later work applies the same idea to smart contracts \citep{luu2016making,bhargavan2016formal} and to purpose-built languages like Catala, which compile statutory text directly into verifiable code \citep{merigoux2021catala}. In each of these systems, a domain expert hand-writes the logic, so correctness is provable but coverage is limited to what has been manually encoded. \tool{} autoformalizes policies, bypassing much of the manual domain-expert labour.

\paragraph{Deontic Reasoning Benchmarks.}
\textsc{ConditionalQA} was the first benchmark to test whether a policy document supports a claim, or that the claim instead depends on conditions not stated \citep{sun2022conditionalqa}. \textsc{RuleBench} and \textsc{RuleArena} generalize this to explicit rule application, spanning domains from content moderation and relation extraction to real-world scenarios like airline baggage fees and tax computation \citep{sun2024rulebench,zhou2025rulearena}. Recently, \textsc{DeonticBench} unifies several such domains (taxes, immigration, housing, airline policy) under a single benchmark \citep{dou2026deonticbench}. These benchmarks establish deontic reasoning as a measurable capability. Each, however, is fixed and manually drawn.

\paragraph{Autoformalization} 
\emph{Autoformalization} is the task of translating natural language into a formal representation \citep{wu2022autoformalization}. It has been studied most extensively for mathematics \citep{jiang2023draft,azerbayev2023proofnet} and formal specifications \citep{cosler2023nl2spec}, and recently for policy documents, where LLMs translate regulatory clauses into logical rules checked by a solver \citep{bayless2025neurosymbolic}. Beyond translation itself, formalization now serves both ends of the LLM pipeline. At test time it acts as a checker: generated code is verified against formal annotations \citep{sun2023clover}, and formalized policies gate model answers at inference \citep{bayless2025neurosymbolic}. At training time it acts as a reward oracle: formally checkable problems supply the verifiable signals behind reasoning-focused reinforcement learning \citep{shao2024deepseekmath,deepseekr1, saturn2025}.

%% file: sections/7-conclusion.tex
\section{Conclusion}

We introduced \tool{}, an automatic, scalable, and parametric benchmark generator for deontic reasoning. \tool{} compiles autoformalized policies into decision trees, encoding potential outcome trajectories for a given verdict. Each path is certified for consistency and entailment, before being added as a benchmark entry. Paths to leaves derive determined benchmark entries, whereas paths to internal nodes do not explicitly entail a verdict, testing model abstention. \tool{} also contains functionality for handling integer datatypes, as well as handling intra-policy dependence.
Across four rule sets, \tool{} generated 263,795 definite and 187,252 \textsf{NEI} entries. On a sample of 1,639 entries, models averaged 75.1--92.5\% correctness on determined verdicts, with a steep decline to 4.0--40.0\% on \textsf{NEI} verdicts.